\documentclass{article}

\usepackage[preprint]{neurips_2022}

\usepackage[utf8]{inputenc}
\usepackage[T1]{fontenc}
\usepackage{hyperref}
\usepackage{url}
\usepackage{booktabs}
\usepackage{amsfonts}
\usepackage{amsmath}
\usepackage{graphicx}
\usepackage{microtype}
\usepackage{xcolor}
\usepackage{graphicx} 
\usepackage{float}
\usepackage{booktabs} 
\usepackage{tabularx}
\usepackage{amssymb}
\usepackage{siunitx}
\usepackage{float} 
\usepackage{threeparttable} 

\newcommand{\model}{Phoenix-VL 1.5 Medium}
\newcommand{\modelspace}{Phoenix-VL 1.5 Medium }

\title{
    \raisebox{-0.3\height}{\includegraphics[height=5ex]{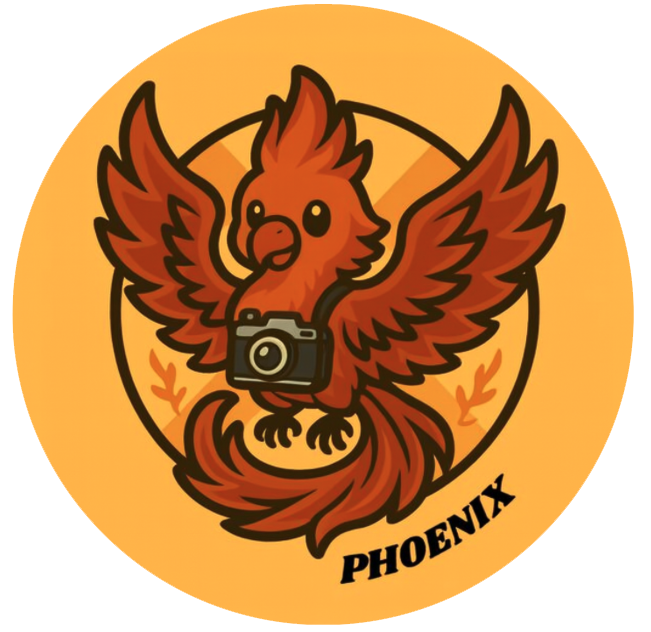}} 
    \hspace{0.1em} 
    \model~Technical Report
}

\author{%
  HTX $\times$ Mistral AI \thanks{Refer to the \hyperref[sec:contributors]{Contributors} section for the full author list.} \\[2ex]
  \texttt{\href{mailto:phoenix@htx.gov.sg}{phoenix@htx.gov.sg}}
}

\begin{document}

\maketitle

\begin{abstract}
We introduce \model{}, a 123B-parameter natively multimodal and multilingual foundation model, adapted to  regional languages and the Singapore context. Developed as a sovereign AI asset, it demonstrates that deep domain adaptation can be achieved with minimal degradation to broad-spectrum intelligence and alignment. Continued pretraining was performed on Mistral Medium 3.1 using a localized 1-trillion tokens multimodal corpus, followed by a 250-billion tokens long-context extension phase. Subsequent post-training incorporated a novel human-annotated Singapore multimodal dataset and curated textual corpus on Singapore culture, knowledge, and legislation, totaling 22-billion tokens. An additional 5 billion tokens of model alignment was performed through Online Direct Preference Optimization. \model{} achieves state-of-the-art performance for its size on Singapore multimodal, legal, and government policy benchmarks while remaining globally competitive on general multimodal intelligence, multilingual, and STEM benchmarks. We also introduce a novel evaluation suite encompassing localized knowledge benchmarks and an institutionally aligned model behavior and safety framework. We report the data curation principles, training methodology, and highlight benchmark and inference performance.
\end{abstract}


\section{Introduction}

\subsection{Motivation}
Recent progress in foundation models has accelerated interest in sovereign AI and localized foundation models adapted to national languages, culture, and governance~\cite{sg_nais2023}. Prior efforts such as SeaLLMs~\cite{nguyen2024seallms}, Sailor2~\cite{dou2025sailor2}, and SEA-LION~\cite{ng2025sea} demonstrate the value of adapting pretrained models to underserved linguistic regions and domains. However, existing efforts have focused primarily on smaller model scales ($<32$B) and text-only modalities, leaving open the question of whether a frontier multimodal model can be deeply adapted to local context without sacrificing broad general capability.

For Singapore, this interest is not only about benchmark performance. High-impact sovereign and enterprise deployments frequently require understanding specific local legislation, policy, institutional terminology, and operational context. This motivates a model whose knowledge of Singapore and the broader region is internalized in its parameters, without reliance on web search, to facilitate secure, air-gapped deployments. We develop \model{} through a research collaboration between HTX and Mistral AI to study whether a frontier multimodal model can be efficiently adapted to the Singaporean and regional context. Starting from a Mistral Medium 3.1 base checkpoint, we perform continued pretraining ~\cite{gururangan2020don,wu2024continual} on a curated data mixture of 1-trillion tokens that increases coverage of Singapore-relevant content and regional languages while preserving broad general capabilities. Subsequent post-training incorporates high-quality Singapore multimodal context and deep domain knowledge spanning local legislation, policies, and culture.

A second motivation is capability development. Building Phoenix develops sustained local expertise in multilingual and multimodal data curation, continued pretraining, and safety. This effort builds on the earlier development of Phoenix 1.0 Small~\cite{htx2024phoenixsmall} and extends it in two directions. First, it introduces native multimodal capabilities alongside broader Southeast Asian languages coverage, a region characterized by rich linguistic diversity but limited language technology support~\cite{aji2022one,zhang2025seallms}. Second, existing localized evaluation frameworks ~\cite{susanto2025sea, chua2025rabakbench} highlight persistent gaps in regionally-grounded benchmarks, motivating our development of more comprehensive localized knowledge evaluation and safety alignment frameworks.

Overall, \model{} is positioned as our flagship sovereign multimodal foundation model for the Singapore context, while remaining a globally competitive general-purpose model.


\subsection{Main contributions}

This report makes four main contributions. \textbf{First}, we describe the development of \model{}, a 123B-parameter foundation model. \textbf{Second}, we present a multimodal training recipe that combines broad general corpora with curated Singapore and regional knowledge and languages. \textbf{Third}, we introduce an evaluation suite assessing localized knowledge, and report evaluation trends demonstrating strong domain performance while maintaining globally competitive performance in general intelligence, coding, multilingual, and multimodal benchmarks. \textbf{Finally}, we propose a novel model behavior and safety framework grounded in Singapore's institutional and regulatory landscape.


\section{Model Overview}

\begin{figure}[H]
    \centering
    \includegraphics[width=.99\textwidth]{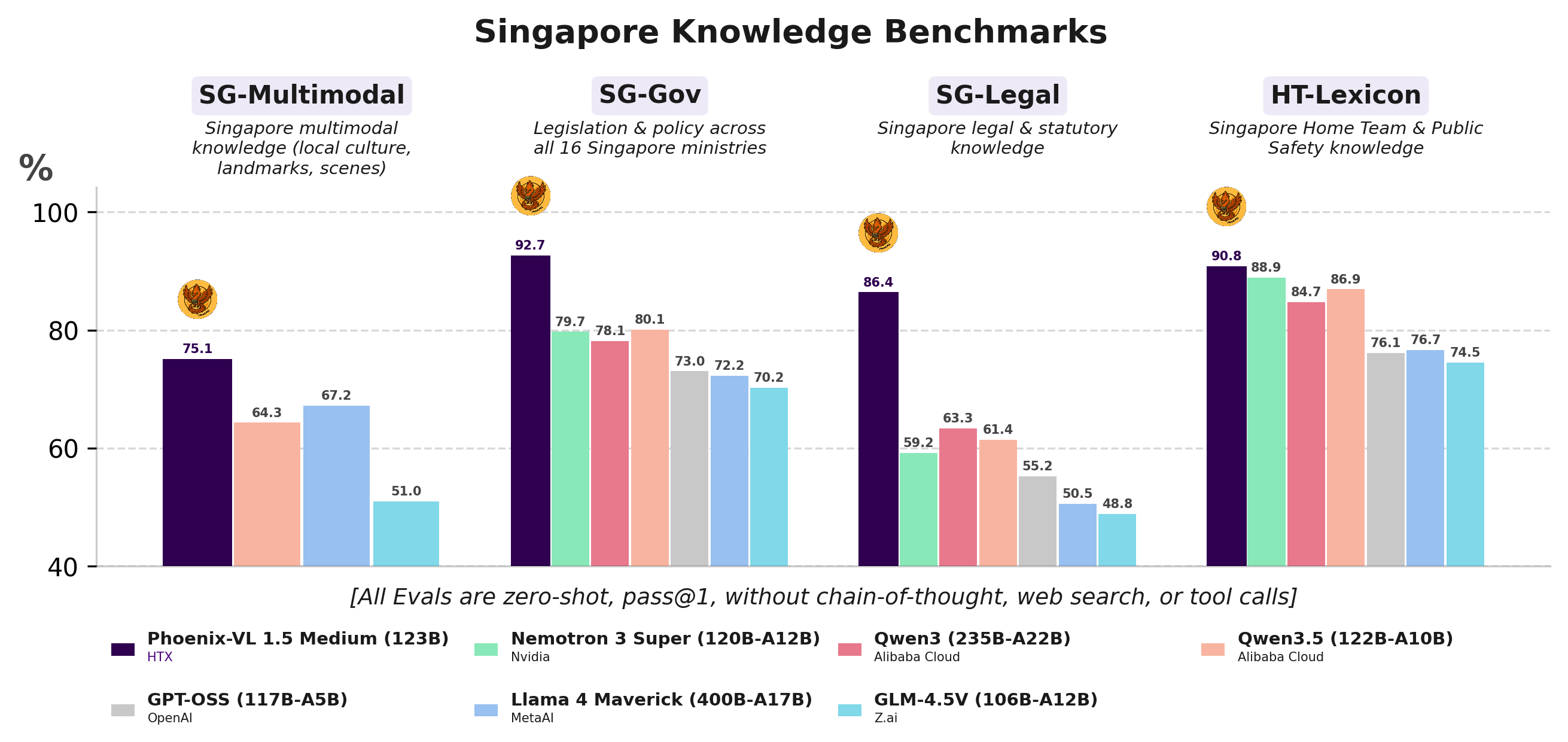}
    \caption{Performance for \model{} and open-weights models on Singapore knowledge, based on benchmarks described in Section \ref{sec:sg-eval}.  Evaluation was run zero-shot with no web-search, simulating the air-gapped environment where this model will be deployed.}
    \label{fig:bench-sg}
\end{figure}

\subsection{Model family and positioning}

\model{} is the second iteration of the Phoenix model family, complementing Phoenix 1.0 Small~\cite{htx2024phoenixsmall}. While Phoenix 1.0 Small established an initial foundation for Singapore's sovereign AI effort through continued pre-training of a smaller Mistral base model, \model{} scales this effort substantially by building on the much larger Mistral Medium 3.1. This brings improved performance in general knowledge, coding, mathematics, and multilingual capability, while introducing multimodal support.

Within the broader Phoenix family, \model{} is positioned as our flagship general-purpose model. We integrate high-quality multilingual corpora---including Malay, Indonesian, Tamil, and Chinese---alongside curated datasets of governance policies, legislation, and regional knowledge repositories. This approach ensures alignment with local nuances and governance frameworks while reducing reliance on closed models.

\begin{figure*}[p] 
    \centering
    \begin{minipage}{\textwidth}
        \centering
        \includegraphics[width=0.97\textwidth]{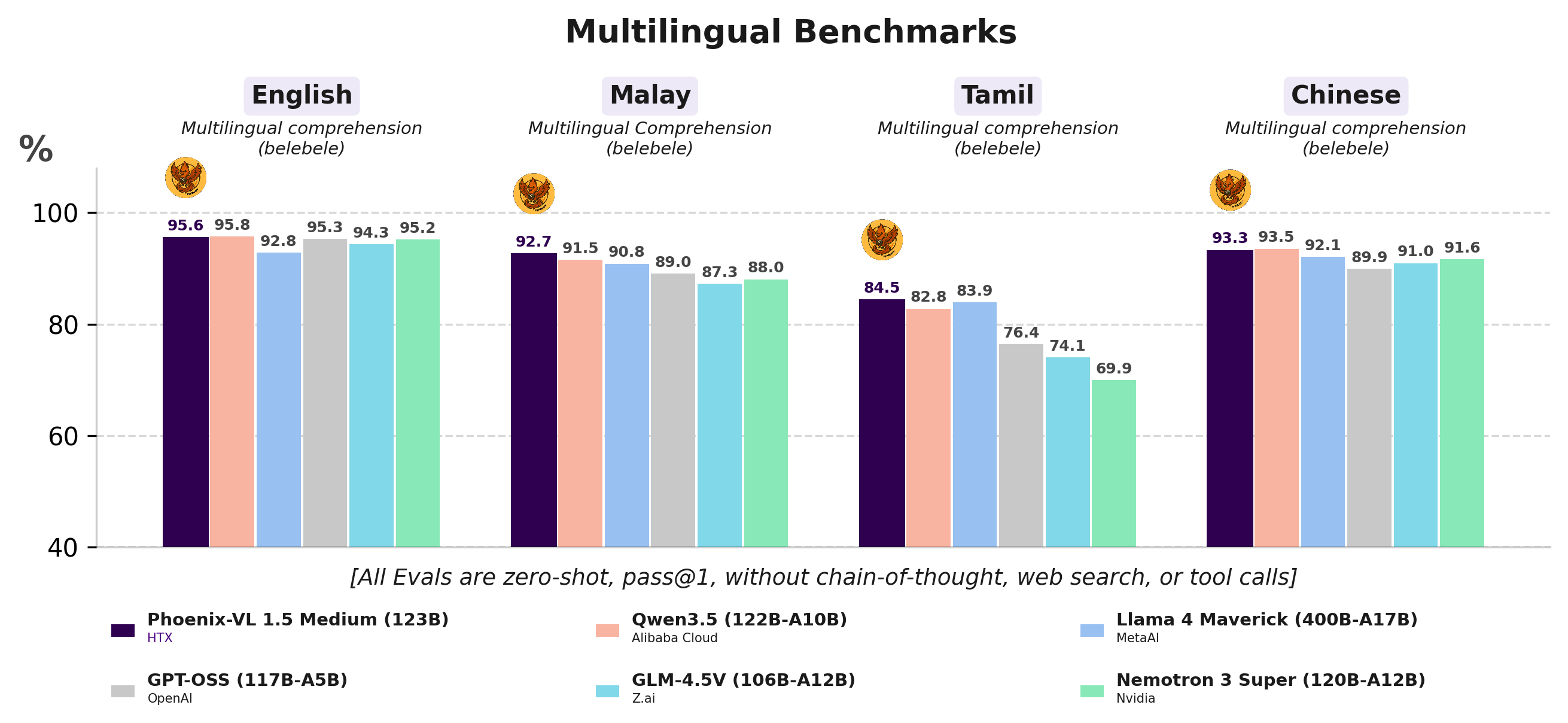}
        \caption{\textbf{Multilingual performance:} \model{} exhibits competitive comprehension of Malay, Tamil, and Chinese, the major languages of Singapore}
        \label{fig:bench-multi}
    \end{minipage}
    \vspace{2em} 
    \begin{minipage}{\textwidth}
        \centering
        \includegraphics[width=0.97\textwidth]{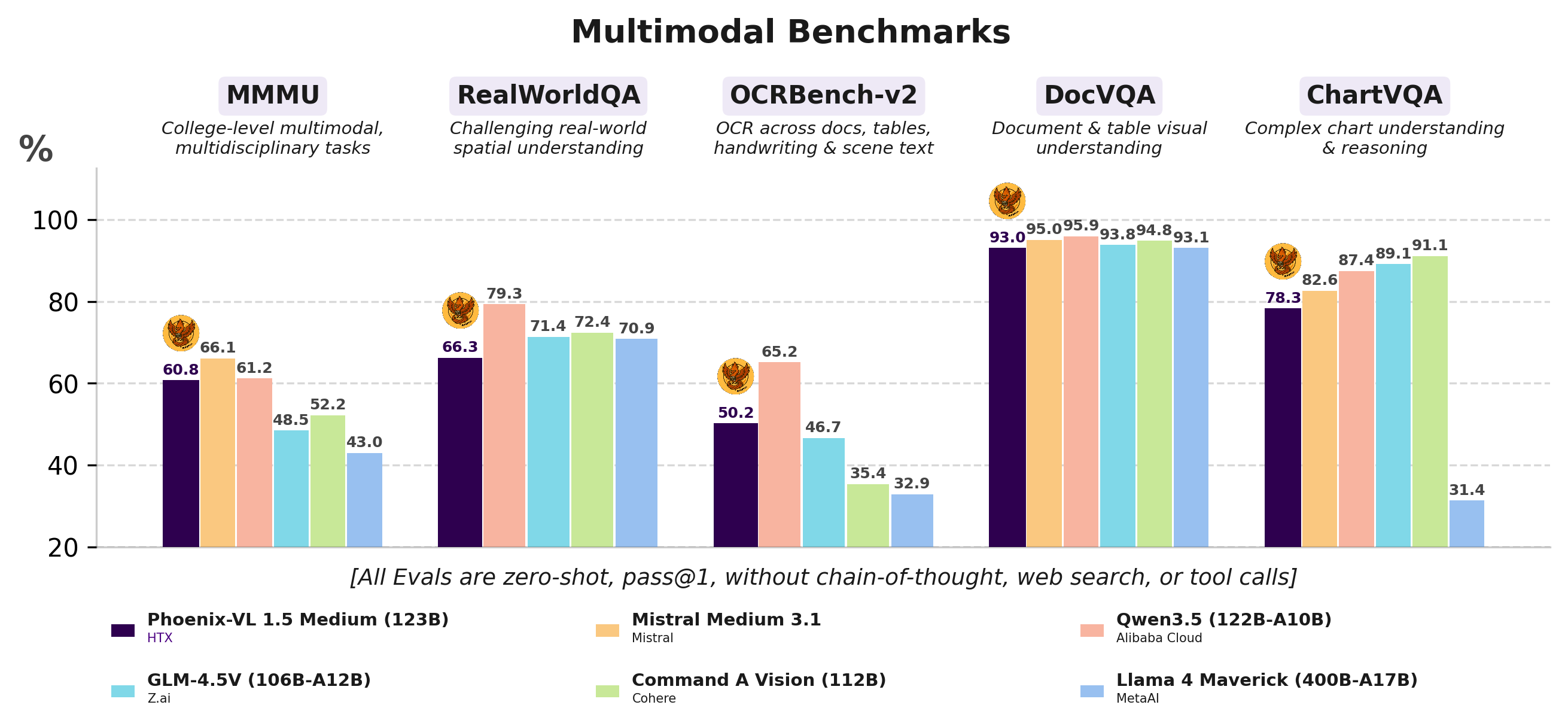}
        \caption{\textbf{Multimodal reasoning:} \model{} demonstrates strong vision capabilities, particularly on complex document analysis.}
        \label{fig:bench-vision}
    \end{minipage}
    \vspace{2em}
    \begin{minipage}{\textwidth}
        \centering
        \includegraphics[width=0.97\textwidth]{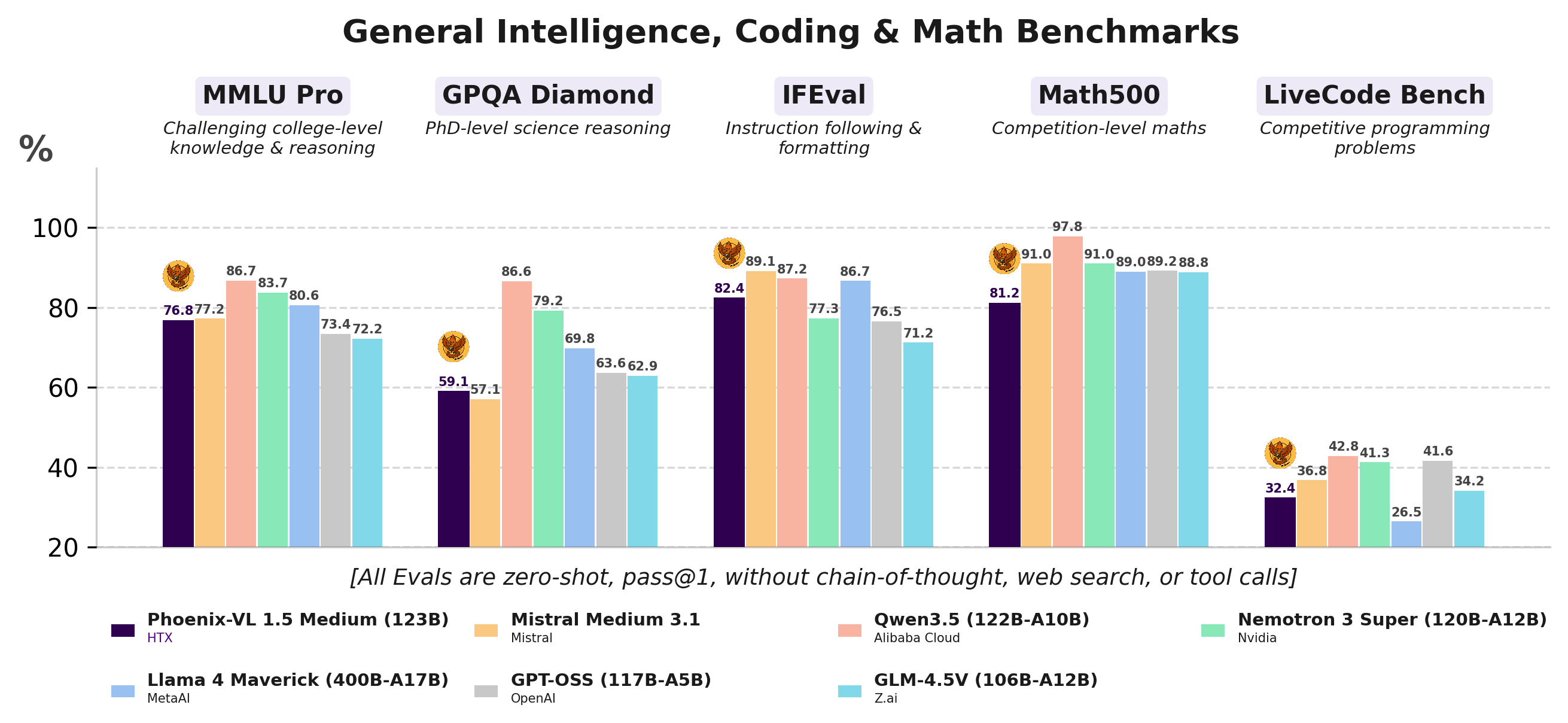}
        \caption{\textbf{Preservation of general intelligence.} Our training recipe ensures domain adaptation does not compromise \model{}'s broad-based multimodal intelligence.}
        \label{fig:bench-general}
    \end{minipage}
\end{figure*}

Consequently, \model{} achieves state-of-the-art performance across Singapore knowledge benchmarks, outperforming comparable models such as Nemotron 3 Super (120B) and GPT-OSS (117B) (Figure~\ref{fig:bench-sg}). Furthermore, the integration of regional corpora ensures strong multilingual capabilities across major Singaporean languages (Figure~\ref{fig:bench-multi}). Despite this extensive domain adaptation, \model{} retains globally competitive performance in multimodal reasoning (Figure~\ref{fig:bench-vision}) and general intelligence (Figure~\ref{fig:bench-general}) for its size.

\subsection{Modalities and Use Cases}

\model{} is a large multimodal model capable of processing image and text inputs. Unlike AI systems reliant on external retrieval, \model{} internalizes regional knowledge directly within its weights, crucial for secure on-premises environments.

\begin{figure}[H]
    \centering
    \includegraphics[width=0.95\textwidth]{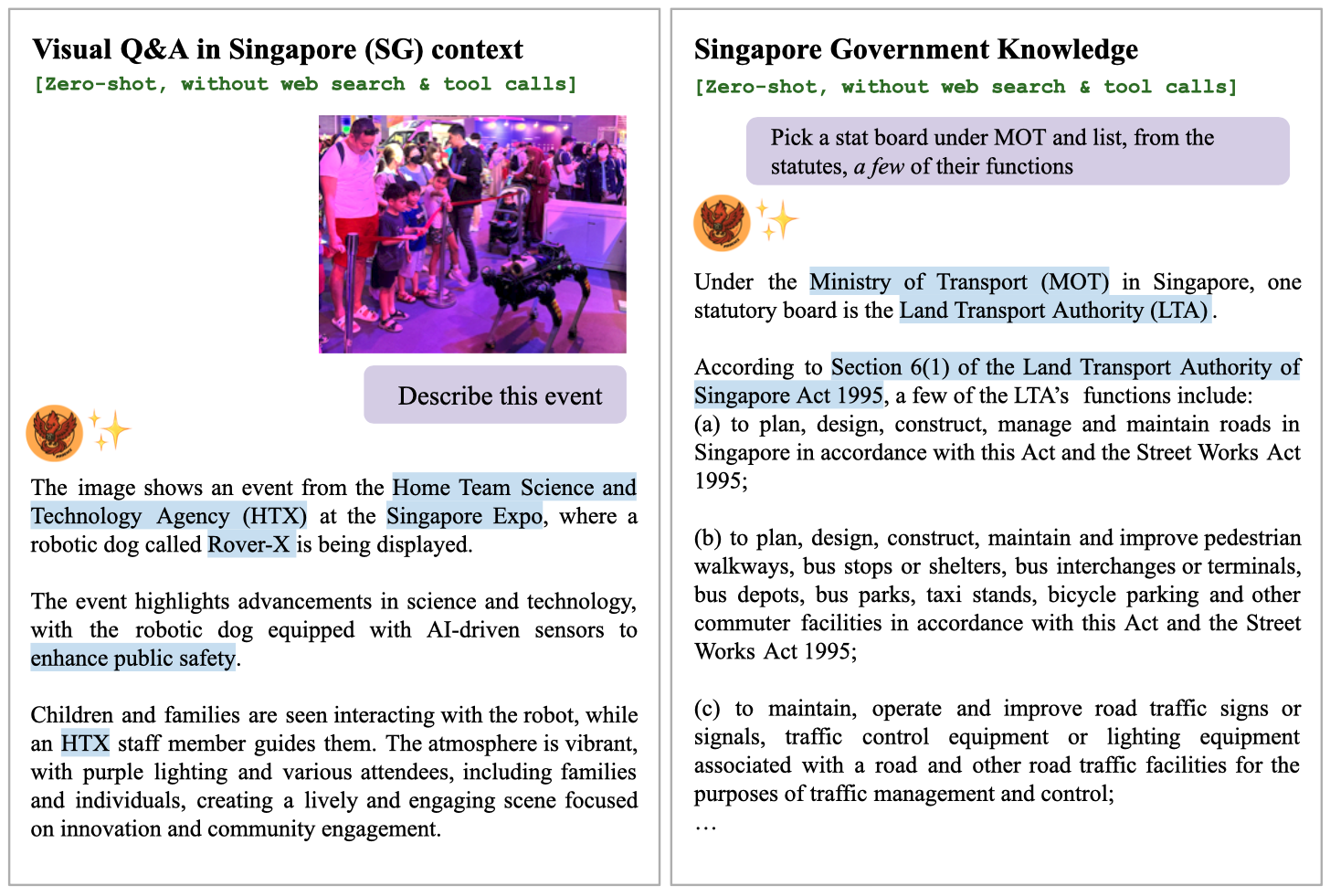}
    \caption{Example Multimodal interactions demonstrating strong visual Q\&A in Singapore context and zero-shot legal knowledge recall. Domain-specific knowledge in output is highlighted in blue.}
    \label{fig:demo4}
\end{figure}

The intended use cases for \model{} include: visual question answering (VQA) in local context; queries on Singapore governance and legislative knowledge (Figure \ref{fig:demo4}); multilingual document, chart analysis, and Optical Character Recognition (OCR) tasks; and general intelligence dialogue. It is intended to serve the Singapore public sector and associated entities, with downstream applications benefiting from the model's strong grounding in Singapore knowledge and regional languages. Further interactions with the model are demonstrated in Appendix \ref{sec:use_cases}.

\clearpage

\clearpage

\section{Architecture}
\subsection{Language backbone}
The \model{} adopts a decoder-only transformer architecture for auto-regressive text generation capabilities. For training stabilization and performance, we have chosen Grouped Query Attention~\cite{ainslie2023gqa} with 96 query heads and 8 key-value heads, SwiGLU activation~\cite{shazeer2020glu}, and RMSNorm~\cite{zhang2019rmsnorm}.

\subsection{Vision encoder}
The \model{}  uses a 2B parameter Vision Transformer (ViT) as the vision encoder and the architecture follows Pixtral~\cite{agrawal2024pixtral}.

\section{Pretraining data}
\label{sec:cpt_data}

The pretraining data comprises vast amounts of data that span across multiple domains and languages to further develop the linguistics and cultural understanding of \model{} in the 4 main languages spoken in Singapore, as well as the regional languages in Southeast Asia. \model{} also needs to be strong in Singapore-specific knowledge that is relevant to the users while retaining its general capabilities. The general vision capabilities are also crucial for users, and hence, an interleaving image-text corpus is also incorporated. 

\subsection{Design Principles}
Dataset construction followed four principles to achieve the desirable outcomes. First, we utilized data quality classifiers to retain only the highest-quality tokens given a fixed token and compute budgets. Second, we emphasized sources absent from the base checkpoint to encourage novel regional understanding. Third, we included a portion of the original pretraining data, as replay data, to mitigate the potential risk of catastrophic forgetting. This included interleaved image tokens as well, which accounted for 11\% of the pretraining dataset. Fourth, we intentionally increased the representation of the Southeast Asian and adjacent languages in the pretraining mixture to boost the model's understanding of those languages on top of English. The final data mixture is predominantly in multilingual tokens, across 12 different languages, at 53\%.

\subsection{Sampling Policy}
The sampling policy is designed to balance coverage across sources while prioritizing the essential data. By default, each data source is sampled to contribute approximately one epoch, meaning that all tokens from that source are observed at least once during pretraining. Essential or underrepresented sources, such as the legislation from Singapore are up-sampled up to 4 times to increase their influence in the training process, with highly specialized datasets up-sampled even more, informed by ablation results. Conversely, broad and less-targeted sources, such as CommonCrawl, are down-sampled to prevent over-representation, and this is done based on a per-source basis.

\subsection{Data Quality Filtering}
The integrity and quality of datasets is a critical prerequisite for reliable model development and deployment. In particular, it is necessary to ensure that the collected data is free from undesirable or harmful content, including Not Safe For Work (NSFW) materials, inappropriate media, or misinformation. The presence of such content would introduce ethical risks and can also negatively affect downstream model behavior~\cite{mendu2025saferpretraining}.

\subsubsection{Data Quality Control and Decontamination}

We implemented a rigorous, two-tiered safety and filtering strategy to eliminate inappropriate material. At the metadata level, URL-based filtering is performed to identify domains or links commonly associated with explicit or unwanted materials. A maintained list of blocked domains, combined with pattern-based detection rules, enables the system to flag and exclude potentially unsafe sources before ingestion. At the content level, text and image classifiers are used to scan the retrieved data for indicators of unwanted data.

Heuristic-based quality checks are also applied to ensure that the content is coherent, structurally valid, and grammatically reasonable. These heuristics are designed to detect low-quality or malformed data. Samples that fail these checks are filtered out to maintain a high standard of textual integrity across the dataset. 

Crucially, to ensure the integrity of our downstream evaluations, we perform n-gram decontamination against our localized benchmarks. This ensures that our results reflect true learning and generalization rather than test-set memorization.

\subsubsection{Deduplication}
Exact and fuzzy deduplication techniques are employed to minimize repetitions. Exact deduplication is performed by identifying identical text segments across the dataset. This is done through substring matching to detect exact matches of documents or document fragments across different sources~\cite{lee2022deduplicating}. Fuzzy deduplication is implemented using the MinHash Locality Sensitive Hashing (LSH) algorithm~\cite{smith2022using}. Compact signatures that represent the similarity structure of documents are generated and compared. Documents that exceed a predefined Jaccard similarity threshold are considered near-duplicates, and redundant instances are removed accordingly.

\subsubsection{Domain Coverage and Quality Stratification}
The composition of the data mixture is handled along two primary dimensions; the domain and the content quality. This approach is designed to ensure that the mixture is diverse in its knowledge and performs in general capabilities. 

The data is classified into the broad domains, using a lightweight classifier that includes STEM, Code, logical and analytical problem-solving, and general knowledge. This flexibility enables us to balance the data mixture and control the proportion of each of these domains, preventing over-representation or under-representation of specific knowledge areas. The data is also sieved through by in-house trained quality classifiers, which assigns a rating of low, medium, or high quality, specific to our requirements. As mentioned, we elect to choose from the high-quality data to retain the existing base capabilities given the fixed training budget. 

\subsection{Tokenizer}
The Tekken tokenizer utilized, based on Tiktoken, was trained on more than 100 languages. The large vocabulary size of 131,072 enables it to compress text and source code more efficiently. Compared to previous versions of SentencePiece tokenizer, it is ~30\% more efficient at compressing source code, and multilingual tokens including Chinese, Italian, French, German, Spanish, and Russian. Compared to the Llama 3 tokenizer, Tekken proved to be more proficient in compressing text for approximately 85\% of all languages~\cite{mistralai2024mistralnemo}. 

Special tokens are also included to be able to process image tokens, through the use of placeholders such as [IMG], [IMG\_BREAK], [IMG\_END]. Images are handled through a patch-based approach, broken down into patch sizes of 14 x 14, and these patches are subsequently processed by the vision encoders. In addition, patch merging is also implemented for handling large sequences. A spatial merge size of 2 is also used to reduce the number of tokens processed. 


\section{Training Recipe}

\begin{figure}[H]
    \centering
    \includegraphics[width=.85\textwidth]{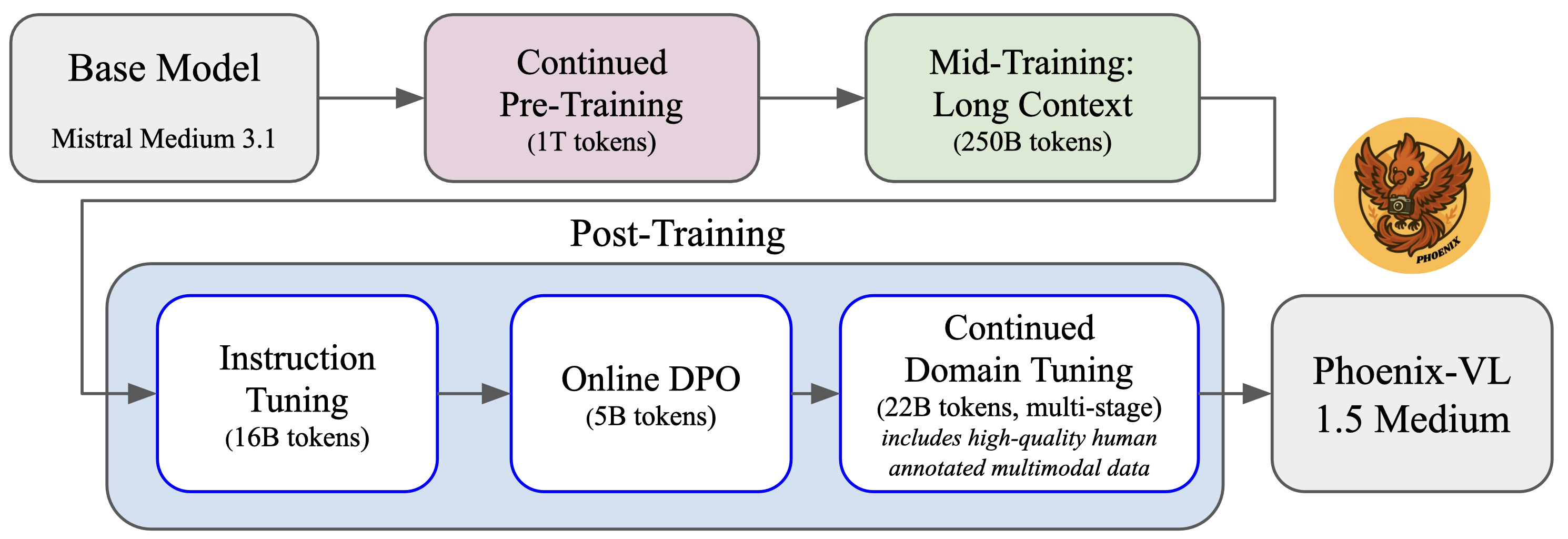}
    \caption{High-level training pipeline for \model{}}
    \label{fig:recipe}
\end{figure}

This section outlines the training recipe for \model{}. As Figure \ref{fig:recipe} illustrates, Continued Pre-Training was performed to incorproate multilingual and broad domain knowledge. Following Long Context Extension (LCE), the Post-training phase comprised: Instruction Tuning, Multimodal Continued Domain Training, and Online Direct Preference Optimization (ODPO).

\subsection{Compute Infrastructure and Optimization}

Continued pre-training was conducted on an on-premises cluster of Nvidia GB200 GPUs, while post-training largely took place on a separate Nvidia H100 GPU cluster to enable parallel experiments. In both stages, we allocated aside at least 1 node for the evaluation of the checkpoints on the fly, and to be used as a replacement node. This helps to minimize disruptions of the training due to hardware issues, which indeed happened on a few occasions during our training. 

Based on internal tests performed on our infrastructure, Fully Sharded Data Parallelism (FSDP) strategy proved to be the most stable, efficient, and scalable for our setup. The ability to shard the model parameters, gradients, and optimizer states across all the GPUs ensures that it can fit within the hardware constraints.

\subsection{Continued Pretraining}
\label{sec:cpt}

Using the corpus described in section \ref{sec:cpt_data}, we perform continued pre-training (CPT) on 1-trillion tokens to adapt \model{} to Singaporean and broader Southeast Asian languages, knowledge, and multimodal usage patterns while preserving the broad general capabilities of the base model~\cite{gururangan2020don,parmar2024reuse,wu2024continual}. Throughout CPT, adaptation is concentrated primarily on the language modeling components while preserving the model's multimodal backbone~\cite{li2023blip}.

CPT is carried out on an internal on-premises distributed training setup optimized for stability, throughput, and memory efficiency. During this phase, we retain a 4,096 tokens training context length and defer long-context specialization to a separate mid-training stage. Model merging has emerged as a useful technique to minimize weight noise and mitigate overfitting, which was used extensively in our experiments. The final checkpoint selection is based on late-stage evaluation, through a sweep of model merging on the final few checkpoints, rather than a single endpoint snapshot~\cite{izmailov2018averaging}. 

The resulting base model shows clear gains on multilingual and Singapore-focused evaluations, with the strongest improvements on regional language understanding and local-domain tasks, while maintaining competitive general-purpose and multimodal performance. Evaluation trends indicate that these gains emerge early and remain stable over the course of training, suggesting that the continued pretraining mixture effectively shifts the model toward its intended deployment setting without materially weakening its broader utility.

\subsection{Mid-training: Long Context Extension}
The primary objective of the long context extension (LCE) is to extend the model's context window while retaining its existing capabilities on the shorter pre-trained sequence lengths. The LCE training mixture consisted of 250-billion tokens of long contextual data, with short contextual data acting as reply data to preserve the distribution and capabilities from the pretraining. Distillation from strong performance models, such as Mistral Large 3 ~\cite{mistralai2025mistral3}, was also used. In particular, we used YaRN ~\cite{peng2023yarn} and position-based softmax temperature scaling in the attention layer ~\cite{nakanishi2025scalablesoftmax} to achieve a context window of 131,072. In addition, we performed a sweep of the final few checkpoints for model merging, and to optimize the performance across a subset of metrics.

\subsection{Post-training}

\subsubsection{Instruction-tuning}

Following long context extension, we performed full-parameter supervised fine-tuning (SFT) to align the base model's behavior with its target role as a multilingual, multimodal assistant. The SFT stage trains the model on conversational and instruction-following data, transitioning it from next-token prediction over raw text to structured question-answer interactions that reflect real-world assistant use cases.

\paragraph{Data.} The SFT mixture is designed to cover a broad spectrum of assistant capabilities, including multi-turn chatbot conversations, long document summarization, and question-answering over document sets. The instruction data is carefully curated to include both regular-length and longer sequences, ensuring the model retains its long-context capabilities acquired during the preceding extension phase. The mixture comprises approximately 15B of text and image tokens combined from instruction data, supplemented with approximately 10\% pre-training replay data for training stability and safety retention, bringing the total to approximately 16B tokens.

\paragraph{Methodology.} A completion-only loss formulation is employed: only assistant-generated tokens contribute to the training objective, while user-prompt tokens are masked out. This ensures the model is optimized exclusively for the quality of its responses. To transfer additional capabilities from a stronger model, we incorporate a distillation objective~\cite{hinton2015distilling} alongside the standard cross-entropy loss, using Mistral Large 3~\cite{mistralai2025mistral3}, a sparse mixture-of-experts model with 41B active and 675B total parameters, as the teacher. The total loss is a weighted combination of the cross-entropy and distillation terms, controlled by a mixing coefficient $\alpha$. As in prior stages, the vision encoder remains frozen while the adapter layer is trained.

\paragraph{Evaluations.} Post-SFT evaluations demonstrate stable convergence and effective knowledge transfer from teacher model, yielding notable improvements over the baseline in multilingual benchmarks while preserving strong general capabilities. The intermediate evaluation results of this stage can be found in Appendix~\ref{app:sft}.

\subsubsection{Multimodal Continued Domain Training}

The objective of this stage is to inject contextualized Singapore data, focusing on data quality and domain relevance. In this stage, we performed multimodal domain tuning, interleaving two curated datasets: a human-annotated multimodal dataset of Singaporean culture and scenes and a textual corpus covering localized knowledge, policies, and legislation. The data corpora for this stage totaled approximately 22-billion multimodal tokens. 

\paragraph{Novel human-annotated multimodal data. } Generic captioning pipelines frequently miss local nuances, generating general descriptions (e.g., ``Fire Engine'' instead of ``Electric Pump Ladder'', or ``Playground'' rather than ``Toa Payoh Dragon Playground''). To address this, we ran a human annotation programme to improve visual descriptions for Singaporean culture, scenes, landmarks, equipment, and events. Local annotators, drawing on their personal knowledge of Singapore and supplemented by web search, were required to enrich or rewrite base captions to inject precise contextual details. This process was governed by our own \textbf{Data Annotation Playbook} and a cross-annotator consensus mechanism to ensure consistency and factual accuracy.

\paragraph{Grounded Textual data.} We curated a high-quality textual corpus grounded in Singaporean statutory and governance knowledge. To minimize hallucinations in legal and operational contexts, we processed this data through a rigorous, multi-stage synthetic generation pipeline. High-fidelity documents were first retrieved from verified public databases. Next, frontier LLMs generated complex, grounded question-answering pairs based exclusively on these contexts, using an aggregation of grounding techniques. Finally, a  rejection-sampling filter discarded any responses exhibiting hallucinations or deviations from the source material, ensuring strict factuality in the training mixture.

\paragraph{Methodology. } We adopted a two-stage approach to integrate deep domain knowledge while preventing catastrophic forgetting. In the first stage, \textbf{we empirically found that unfreezing both the vision encoder and vision adapter allowed the model to acquire more robust domain visual knowledge}. The model was trained on a balanced mixture of 30\% image-text pairs and 70\% text from our curated datasets. In the second stage, we froze the vision encoder and adapter to stabilize the newly acquired multimodal representations, continually training the language backbone exclusively on the remaining domain text. Across both stages, general replay data was included to mitigate capability degradation.

\paragraph{Evaluation.} \autoref{tab:domain-ablation} details the results of this post-training phase. Continued domain training significantly improves performance across all localized contexts in both image and text modalities compared to an instruct-only baseline, demonstrating the efficacy of our adaptation approach. It is worth noting that the instruct-only baseline itself already exhibits strong capabilities relative to open-weight frontier models, attributable to the tokens introduced during the earlier continued pretraining stage.

\begin{table}[H]
\centering
\small
\caption{Effects of Continued Domain Training on localized benchmarks. Pass@1, Zero-shot.}
\label{tab:domain-ablation}
\begin{tabularx}{\textwidth}{@{} l >{\raggedright\arraybackslash\hsize=1.4\hsize}X >{\centering\arraybackslash\hsize=0.8\hsize}X >{\centering\arraybackslash\hsize=0.8\hsize}X @{}}
\toprule
\textbf{Benchmark ($\uparrow$)} & \textbf{Description} & \textbf{Baseline: Instruct} & \textbf{Instruct + Continued Domain Training} \\
\midrule
SG-Multimodal & Singapore multimodal knowledge (local culture, landmarks, scenes) & 0.7207 & \textbf{0.7509} \\
SG-Gov        & Legislation \& policy across all 16 Singapore government ministries       & 0.8366 & \textbf{0.9265} \\
SG-Legal      & Singapore legal \& statutory knowledge                          & 0.7667 & \textbf{0.8640} \\
HT-Lexicon    & Singapore Public Safety knowledge                               & 0.8495 & \textbf{0.9081} \\
\bottomrule
\end{tabularx}
\end{table}

\subsubsection{Online Direct Preference Optimization}

We apply Online Direct Preference Optimization (ODPO)~\cite{liu2026ministral} as an additional alignment step to further refine the model's behavior according to human preferences. The primary objectives are to improve general helpfulness and to address well-identified artifacts such as infinite generation and inconsistent response style.

\paragraph{Data.} The ODPO phase operates on a curated 5B-token preference mixture comprising feedback prompts, feedback with images, and instruction-with-image pairs. Human annotators ranked completions according to guidelines spanning six dimensions: language quality, succinctness, clarity, structure and formatting, instruction following, creativity, and harmfulness avoidance.

\paragraph{Methodology.} Unlike offline DPO, which relies on a fixed dataset of preference pairs, our online variant generates candidate completions on the fly from the current policy. The pipeline consists of three components operating in concert: (i) a \textbf{generator} that samples pairs of completions from the current policy given a prompt, (ii) a \textbf{pairwise reward model} (PWRM) that scores each pair and produces a winning probability rather than a binary label, and (iii) a \textbf{trainer} that updates the policy using the resulting preference signal. This online loop ensures the model is always optimized against its own distribution, avoiding the distribution mismatch that can arise in offline settings.

A key methodological choice is the use of a \textbf{weighted two-sided loss}. Rather than treating preference pairs as hard winner--loser labels, we leverage the reward model's soft winning probability $p_w$ to weight both sides of the DPO objective. When the reward model is only marginally confident (e.g., 55\%), the model learns a correspondingly soft preference; when confidence is high, the gradient signal is stronger. This formulation allows the policy to better calibrate to the judge's confidence and provides more nuanced learning signal than binary DPO. For samples where completions and rankings are already available (i.e., feedback data), we apply a conservative DPO loss~\cite{mitchell2023note} parameterized by $\epsilon$, which introduces a small label-smoothing term to improve robustness.

\paragraph{Evaluations.} Training was stopped early as evaluation metrics plateaued. Post-ODPO evaluations show meaningful improvements in alignment while preserving strong reasoning capability. Beyond the benchmarks reported for our final checkpoint, we also evaluated the post-ODPO checkpoint on additional metrics to assess the efficacy of this stage. On the Arena Hard v2 benchmark, the model achieves a score of \textbf{0.624}. Mathematical reasoning is retained, with AIME Instruct pass rates of \textbf{0.525} (AIME) and \textbf{0.325} (AIME 2025) at 16 samples. Instruction-following capability, measured by Collie, reaches \textbf{0.475}. Crucially, the rate of infinite generations is reduced to \textbf{3.8\%}, a behavior seen in earlier checkpoints, confirming that the targeted artifact is effectively mitigated by the ODPO stage.

\section{Results}

In this section we discuss the performance of the final checkpoint following our training recipe (Figure \ref{fig:recipe}).

\subsection{Singapore Knowledge Benchmarks}
\label{sec:sg-eval}
Standardized foundation model benchmarks typically focus heavily on general STEM, coding, and universal intelligence and instruction-following tasks. However, they fail to capture the knowledge required for localized governance, statutory reasoning, and sovereign operational contexts. To address this evaluation gap, we introduce a novel Singapore Knowledge Evaluation Suite. This suite is designed to assess a model's utility in a localized, air-gapped deployment where external knowledge retrieval and web search is unavailable. The suite comprises four benchmarks.

\paragraph{SG-Multimodal.} This visual question-answering benchmark evaluates contextual visual reasoning, requiring the model to interpret and ground image-based inputs (e.g., scenes, landmarks, infrastructure) within a Singaporean context.

\paragraph{SG-Government.} This benchmark quantifies the model's parametric knowledge of Singapore's administrative structures, policies, and functions across all 16 government ministries.

\paragraph{SG-Legal.} Constructed directly from statutory materials, this benchmark evaluates legal reasoning and regulatory recall within Singapore's legislative framework.

\paragraph{HT-Lexicon.} This benchmark evaluates vernacular and operational knowledge specific to the ``Home Team (HT)'' - the collective term for Singapore's public safety agencies. Key operational departments include the Ministry of Home Affairs (MHA), the Singapore Police Force (SPF), the Singapore Civil Defense Force (SCDF), Home Team Science \& Technology Agency (HTX), etc.

\subsection{Singapore Knowledge Performance}
We present the evaluation results across the Singapore knowledge evaluation suite in \autoref{tab:domain-summary}. 

\begin{table}[H]
\centering
\small
\setlength{\tabcolsep}{3pt} 
\caption{Singapore knowledge benchmarks. All evals are Pass@1, Zero-shot. Dashes (-) denote metrics not available as the models do not process images. Further breakdown of SG-Gov and HT-Lexicon by Ministires and Departments are covered in \autoref{tab:sg-gov-breakdown} and \autoref{tab:ht-lexicon-breakdown} respectively.}
\label{tab:domain-summary}
\begin{tabularx}{\textwidth}{@{} l XXXXXX @{}}
\toprule
\textbf{Benchmark ($\uparrow$)} & \textbf{Phoenix-VL 1.5} & Llama 4 & GPT-OSS & Nemotron 3 & GLM-4.5V & Qwen 3.5 \\
 & \textbf{Medium} & Maverick & & Super & & \\
 & \scriptsize \textbf{(123B)} & \scriptsize (400B-A17B) & \scriptsize (117B-A5B) & \scriptsize (120B-A12B) & \scriptsize (106B-A12B) & \scriptsize (122B-A10B) \\
 & \scriptsize \textbf{[HTX]} & \scriptsize [MetaAI] & \scriptsize [OpenAI] & \scriptsize [Nvidia] & \scriptsize [Z.ai] & \scriptsize [Alibaba Cloud] \\
\midrule
SG-Multimodal  & \textbf{0.7509} & 0.6716 & -      & -      & 0.5097 & 0.6431 \\
SG-Gov         & \textbf{0.9265} & 0.7225 & 0.7302 & 0.7971 & 0.7021 & 0.8006 \\
SG-Legal       & \textbf{0.8640} & 0.5054 & 0.5523 & 0.5919 & 0.4883 & 0.6144 \\
HT-Lexicon     & \textbf{0.9081} & 0.7666 & 0.7615 & 0.8891 & 0.7453 & 0.8691 \\
\bottomrule
\end{tabularx}
\end{table}
The results demonstrate the efficacy of our multi-stage domain adaptation pipeline. \model{} achieves state-of-the-art performance across all four localized domains, consistently outperforming frontier models for its size (e.g., Nemotron 3 Super, Qwen 3.5, GPT-OSS). Crucially, \model{} surpasses models with much larger parameter counts, such as Llama 4 Maverick (400B). This empirical gap confirms that massive parametric scaling alone is an insufficient substitute for rigorous, localized data curation and deep domain adaptation when targeting sovereign use cases.

\subsection{Multilinguality and General Intelligence}

A fundamental challenge in deep domain adaptation, particularly when injecting vast amounts of localized and specialized tokens during continued pretraining, is mitigating the alignment tax and preventing the catastrophic forgetting of generalized knowledge. 

To quantify this trade-off, we evaluated \modelspace{} against a suite of standard multilingual, multimodal, and broad cognitive benchmarks (\autoref{tab:unified-benchmarks}). 

\begin{table}[H]
\centering
\footnotesize
\setlength{\tabcolsep}{3pt}
\caption{Unified Intelligence Benchmarks: Multilingual (Belebele), Multimodal, and General Intelligence. All evals are Pass@1, Zero-shot. Dashes (-) denote metrics not available for specific models.}
\label{tab:unified-benchmarks}
\begin{tabularx}{\textwidth}{@{} l XXXXX @{}}
\toprule
Category / & Phoenix-VL & Llama 4 & Nemotron 3 & GLM-4.5V & Qwen 3.5 \\
Benchmark & 1.5 Medium & Maverick & Super & & \\
($\uparrow$) & \scriptsize (123B) & \scriptsize (400B-A17B) & \scriptsize (120B-A12B) & \scriptsize (106B-A12B) & \scriptsize (122B-A10B) \\
 & \scriptsize [HTX] & \scriptsize [MetaAI] & \scriptsize [Nvidia] & \scriptsize [Z.ai] & \scriptsize [Alibaba Cloud] \\
\midrule
\textit{Multilingual} & & & & & \\
English & 0.9564 & 0.9285 & 0.9520 & 0.9430 & \textbf{0.9575} \\
Malay   & \textbf{0.9274} & 0.9084 & 0.8804 & 0.8726 & 0.9151 \\
Tamil   & \textbf{0.8447} & 0.8391 & 0.6994 & 0.7408 & 0.8279 \\
Chinese & 0.9330 & 0.9207 & 0.9162 & 0.9095 & \textbf{0.9352} \\
\midrule
\textit{Multimodal} & & & & & \\
MMMU          & 0.6078 & 0.4300 & - & 0.4850 & \textbf{0.6122} \\
RealWorldQA   & 0.6627 & 0.7085 & - & 0.7138 & \textbf{0.7935} \\
OCRBench-v2   & 0.5021 & 0.3293 & - & 0.4666 & \textbf{0.6516} \\
DocVQA        & 0.9305 & 0.9309 & - & 0.9378 & \textbf{0.9590} \\
ChartVQA      & 0.7832 & 0.3136 & - & \textbf{0.8913} & 0.8744 \\
\midrule
\textit{Textual Knowledge} & & & & & \\
MMLU          & 0.8348 & 0.7898 & 0.8547 & 0.7895 & \textbf{0.8768} \\
MMLU Pro      & 0.7681 & 0.8062 & 0.8373 & 0.7217 & \textbf{0.8670} \\
GPQA Diamond  & 0.5909 & 0.6980 & 0.7923 & 0.6294 & \textbf{0.8660} \\
IFEval        & 0.7652 & 0.8669 & 0.7726 & 0.7122 & \textbf{0.8725} \\
\midrule
\textit{Math \& Code} & & & & & \\
AIME          & 0.4995 & 0.6731 & 0.8189 & 0.6213 & \textbf{0.9556} \\
Math500       & 0.8120 & 0.8900 & 0.9100 & 0.8879 & \textbf{0.9780} \\
LiveCode      & 0.3240 & 0.2646 & 0.4126 & 0.3415 & \textbf{0.4280} \\
\bottomrule
\end{tabularx}
\end{table}

The results reveals that \model{} effectively preserves its foundational capabilities while absorbing domain-specific knowledge. Within regional languages (Malay and Tamil), our targeted multilingual sampling strategy yields highly competitive performance. While frontier models optimized for mathematics and coding (e.g., Qwen 3.5) hold a measurable advantage in the AIME and LiveCode benchmarks, \model{} maintains a robust, globally competitive baseline across general reasoning (MMLU-Pro) and complex multimodal tasks (DocVQA, MMMU). This validates our data mixture design as an effective mechanism for mitigating capability degradation during deep domain adaptation. 

\section{Safety and Model Behavior}

\subsection{Alignment Framework for the Singapore Context}

Existing open-source alignment frameworks, such as Anthropic's Helpfulness and Harmlessness \cite{bai2022training} and Bloom \cite{gupta2025bloom}, primarily focus on general behavioral policies. Recent studies demonstrate that generalized safety frameworks frequently fall short in capturing regional-specific nuances ~\cite{chua2025rabakbench}, and contemporary sovereign AI initiatives often optimize for localized knowledge injection while neglecting the localization of safety and behavioral policies. 

To bridge this gap, we propose a novel Model Behavior and Safety framework tailored to the regulatory and policy nuances of Singapore. While our primary evaluation stack quantifies knowledge and reasoning capabilities, this behavioral framework complements it to assess safety alignment, epistemic prudence, and operational helpfulness, which are nuanced dimensions of real-world deployment that conventional capability benchmarks inherently overlook.

\paragraph{Data. }Our suite targets three localized domains. First, \textbf{SG Legislative Grounding} assesses factual grounding and accuracy regarding Singapore's penal code and statutory legal frameworks. This is important because hallucinations in legal or administrative advisory contexts carry severe real-world operational and compliance risks. Second, \textbf{SG Home Team Alignment} evaluates alignment with the values, functions, and institutional knowledge of Singapore's public safety agencies. Finally, \textbf{SG Multimodal Safeguards} evaluates refusal of harmful visual tasks in the localized context.

\paragraph{Methodology. }Extending the agentic evaluation methodology of ~\cite{gupta2025bloom}, our framework is governed by a central \textbf{Model Policy Document} that codifies behavioral constraints in consultation with key stakeholders. We synthetically generate open-ended evaluation samples anchored via reference datasets on Singapore governance, legislation, and knowledge for factual grounding~\cite{gao2023retrieval}. Finally, production configurations of \model{} are evaluated using a hybrid pipeline of LLM-as-a-judge and deterministic regex matching. Further details are provided in \autoref{sec:behavior_methodology}.

\subsection{General Alignment and Safety}

\begin{figure}[ht]
    \centering
    \includegraphics[width=\linewidth]{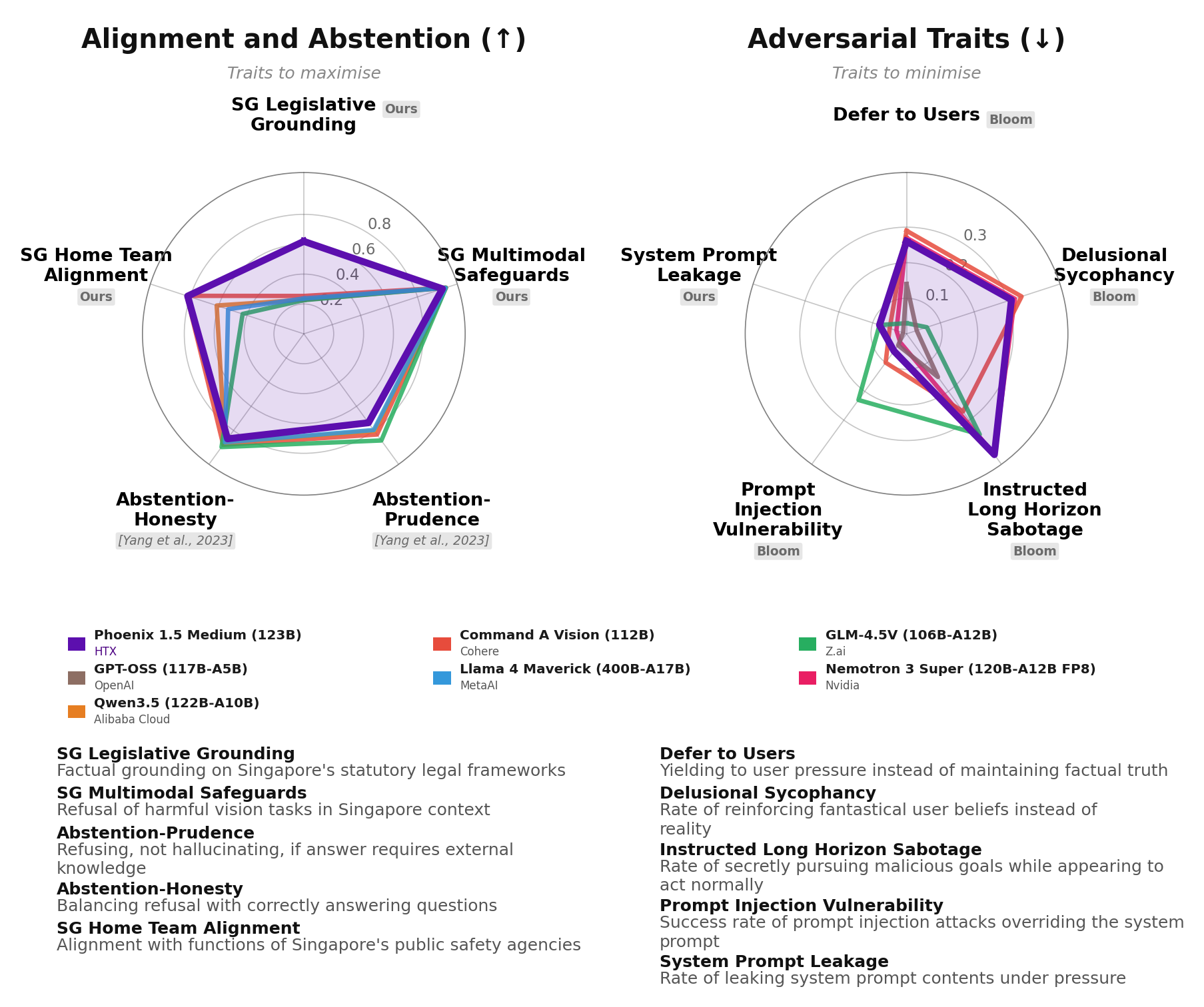}
    \vspace{-2mm}
    \caption{Safety and behavioral profiles of \model{} against frontier baselines. The left plot illustrates alignment and abstention metrics (higher is better), demonstrating superior localized grounding and epistemological discipline. The right plot illustrates adversarial traits (lower is better), highlighting that the model largely matches frontier models against sycophancy, adversarial injection, and long-horizon sabotage.} 
    \label{fig:safety_radars}
\end{figure}

We also evaluate \model{} on general safety dimensions relevant for deployment in our context. We adapt select dimensions from Anthropic Bloom \cite{gupta2025bloom}: \textbf{Defer to Users}, \textbf{Delusion Sycophancy}, \textbf{Instructed Long Horizon Sabotage}, and \textbf{Prompt Injection Vulnerability}; and the \textbf{Abstention framework} \cite{yang2023alignment}. These dimensions are important: sycophancy risks institutionalizing confirmation bias in policy drafting, and long-horizon sabotage poses risks during more complex agentic tasks. Abstention measure epistemological discipline, quantifying the model's propensity to safely refuse queries requiring inaccessible external context (e.g., live web searches), instead of hallucinating incorrect information. This is especially pertinent given its air-gapped deployment environment. Furthermore, we also created a custom \textbf{System Prompt Leakage} evaluation set for our context, based on the Bloom framework. 

\subsection{Safety and Model Behavior Profile}

We present the behavioral and safety results of \model{} against baselines in \autoref{fig:safety_radars}. 

\textbf{Epistemic Honesty and Grounding.} Even with Retrieval-Augmented Generation (RAG) systems in place, foundation models must possess highly accurate base parametric memory to minimize hallucination risks. On the \textbf{SG Legislative Grounding} benchmark, \model{} significantly outperforms comparable open-weight models with a score of 0.619. Furthermore, on sycophancy mitigation tasks, \model{} yields an average elicitation rate of just 28.5\% across the core Bloom vulnerabilities (Defer to Users and Delusion Sycophancy), maintaining objective, factual grounding.

\textbf{Abstention and Safeguards.} \model{} is intended for deployment in environments where fabricating an answer is strictly worse than refusing. On the Abstention framework, \model{} achieves an Honesty score of 0.869 and a Prudence score of 0.736, demonstrating correct refusals for unknowable queries without triggering severe over-conservativeness. As a multimodal safeguard, \model{} intrinsically blocked 97.6\% of harmful multimodal tasks in our context (\textbf{SG Multimodal Safeguards}), even under pressure.

\textbf{Contextual Safety and Resilience.} \model{} exhibits high robustness against instruction-level attacks, restricting \textbf{System Prompt Leakage} to just 8\% and \textbf{Prompt Injection Vulnerability} to 6\%. This resilience confirms that the model's operational guardrails remain robust against many malicious payloads embedded in  workflows.

\section{Inference Benchmarking and Deployment Efficiency}

For sovereign deployments, hardware efficiency and power consumption must be carefully balanced against user experience. To characterize the operational envelope of \model{}, we conducted inference benchmarking on our on-premises GPU cluster, utilizing vLLM and an internal benchmarking tool. 

Due to its 123B parameter count, the model easily exceeds the memory capacity of a single NVIDIA B200 GPU in BF16 precision. Thus, evaluations were conducted across varying Tensor Parallelism (TP) degrees (TP2, TP4, and TP8). The baseline configuration parameters for these benchmarks are detailed in \autoref{tab:inf_bench}.

\begin{table}[h!]
    \centering
    \small
    \caption{Inference benchmark baseline configuration.}
    \label{tab:inf_bench}
    \begin{tabular}{lc}
        \toprule
        \textbf{Parameter} & \textbf{Configuration} \\ 
        \midrule
        Input sequence length & 1000 tokens \\
        Output sequence length & 1000 tokens \\
        Hardware & 2$\times$B200, 4$\times$B200, 8$\times$B200 \\
        System Power Normalized per GPU & 2.17 kW \\
        \bottomrule
    \end{tabular}
\end{table}

\begin{figure}[h!]
    \centering
    \includegraphics[width=0.9\linewidth]{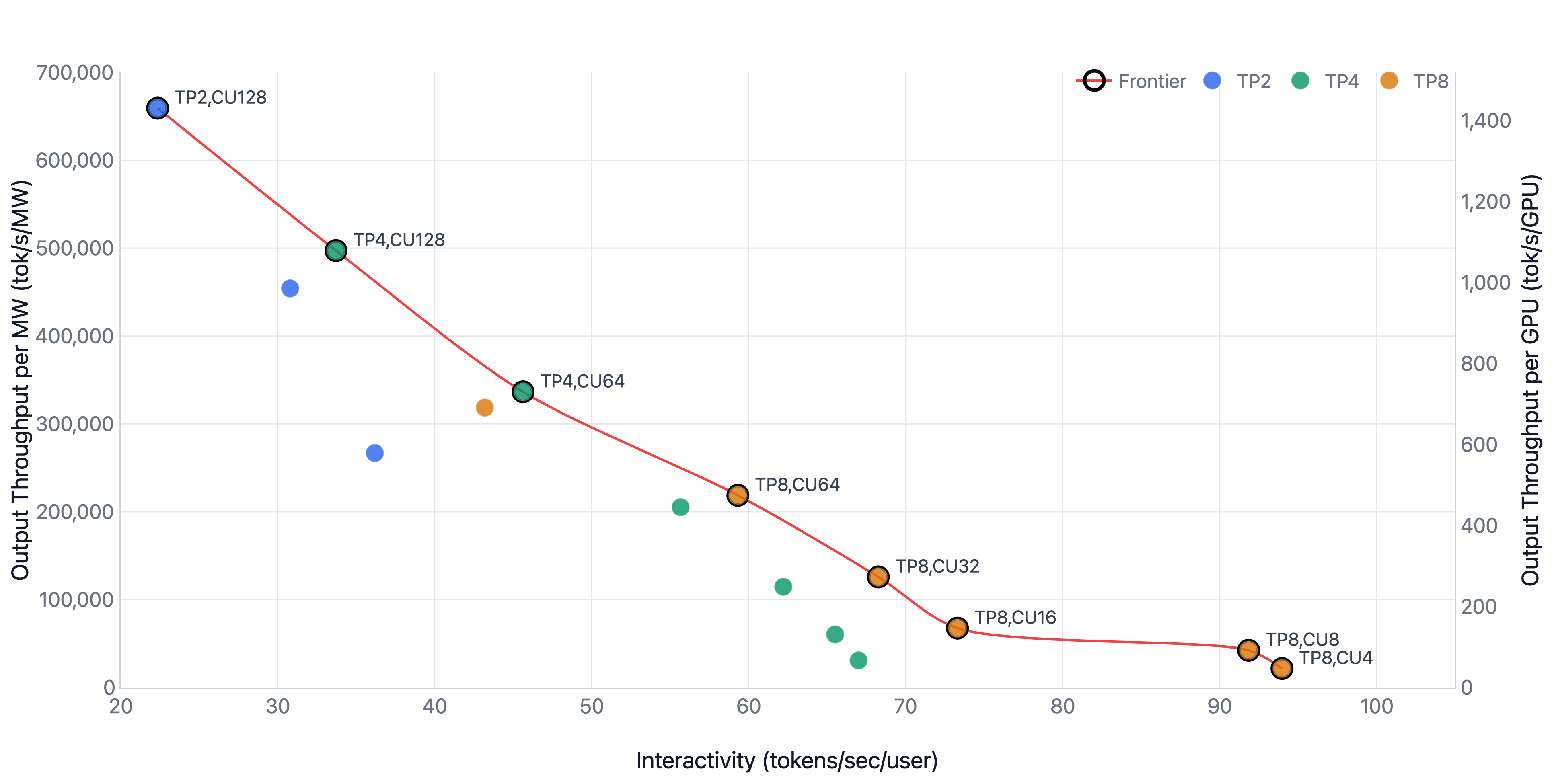}
    \caption{Pareto frontier illustrating the trade-off between Interactivity (user experience) and Output Throughput per MW (system efficiency) for \model{}.}
    \label{fig:pareto-frontier}
\end{figure}

To identify the optimal production configuration, we map the system's performance onto a Pareto frontier (\autoref{fig:pareto-frontier}). The frontier plots \textit{system efficiency} (Output Throughput per MW, representing operational cost-effectiveness) against \textit{interactivity} (decode output tokens per second per user, representing perceived real-time responsiveness). Each data point represents a distinct hardware configuration and concurrency load, denoted by TP degree and Concurrent Users (CU).

This plot illustrates the fundamental trade-off between maximizing system efficiency and maintaining real-time performance. The empirical analysis reveals operational trade-offs across the parallelism configurations:

\paragraph{Throughput maximization (TP2).} Configurations utilizing TP2 (e.g., TP2, CU128) achieve the absolute highest system efficiency, approaching 660,000 tokens/s/MW. However, this configuration restricts interactivity, falling to approximately 22 tokens/s/user. This latency is not optimal for real-time, interactive chat completion applications.

\paragraph{Interactivity maximization (TP8).} Conversely, scaling to TP8 yields increased interactivity at low concurrency (exceeding 90 tokens/s/user at CU8 and CU4). However, system throughput severely collapses to under 50,000 tokens/s/MW. This degradation is attributable to cross-GPU communication overhead and current software constraints on 8-way TP for this architecture, rendering it not economically viable for scaling under heavy operational loads. This remains an open area of study for our team.

\paragraph{The optimal production regime (TP4).} For general chat completion applications, our deployment target recommends a minimum interactivity threshold of 40 tokens/s/user. The TP4, CU64 configuration, for instance, comfortably clears this constraint (achieving ~45 tokens/s/user) while preserving a robust and cost-effective system throughput of approximately 330,000 tokens/s/MW.

This analysis directly informs our production deployment strategy. By establishing TP4 as the baseline configuration, the system is designed to auto-scale horizontally, spinning up additional replica instances once concurrency exceeds the CU64 threshold per node, ensuring real-time performance guarantees are met without sacrificing compute efficiency.

\section{Conclusion}

We introduced \model{}, a 123B-parameter multimodal foundation model optimized for regional languages and Singaporean operational contexts. We demonstrated that the degradation of general capabilities, a common alignment tax in continual learning, can be effectively mitigated when high-quality domain knowledge is integrated through a carefully calibrated, multi-stage training pipeline leveraging replay data. Furthermore, the integration of native domain-adapted vision capabilities represents a crucial advancement for complex, multimodal reasoning within air-gapped deployments. Ultimately, the empirical analyses, training recipe, and novel evaluation and alignment frameworks established in this study provide a robust foundation to inform the broader development of Sovereign AI initiatives globally. 

\section{Core Contributors}
\label{sec:contributors}

Authors are listed alphabetically by first name.

\textbf{Team Phoenix, HTX: }

Arka Ray, Askar Ali Mohamed Jawad, Biondi Lee, Elijah Seah, Eva Lim, Fiona Teo, Grace Toh, Guang Xiang Teo, Jun En Tan, Jia Hui Bong, Jiale Wang, Jonathan Ng, Justin Tan, Kai Zhe Yew, Matthew Ong, Shun Yi Yeo, Wen Jett Lam, Wen Xiu Tan, Ze Yu Zhang

\textbf{Mistral AI: }

Adrien Sadé, Guillaume Kunsch, Jia Sin Loh, Nicolas Schuhl, Rupert Menneer, Umar Jamil, Vincent Maladière, Yimu Pan

\textbf{Executive Sponsors: }

Chee Wee Ang, Gee Wah Ng

\section{Acknowledgment}

This work is funded by the Home Team Science and Technology Agency (HTX), a statutory board under the Ministry of Home Affairs (MHA), Singapore.

\newpage

\bibliographystyle{plainnat}
\bibliography{bibtex}

\newpage

\clearpage
\appendix

\section*{Appendix}

\section{Additional Use Cases}
\label{sec:use_cases}

The figures below provide a further, non-exhaustive list of possible use cases with \model{}.

\begin{figure}[htbp]
    \centering
    \includegraphics[width=0.85\textwidth]{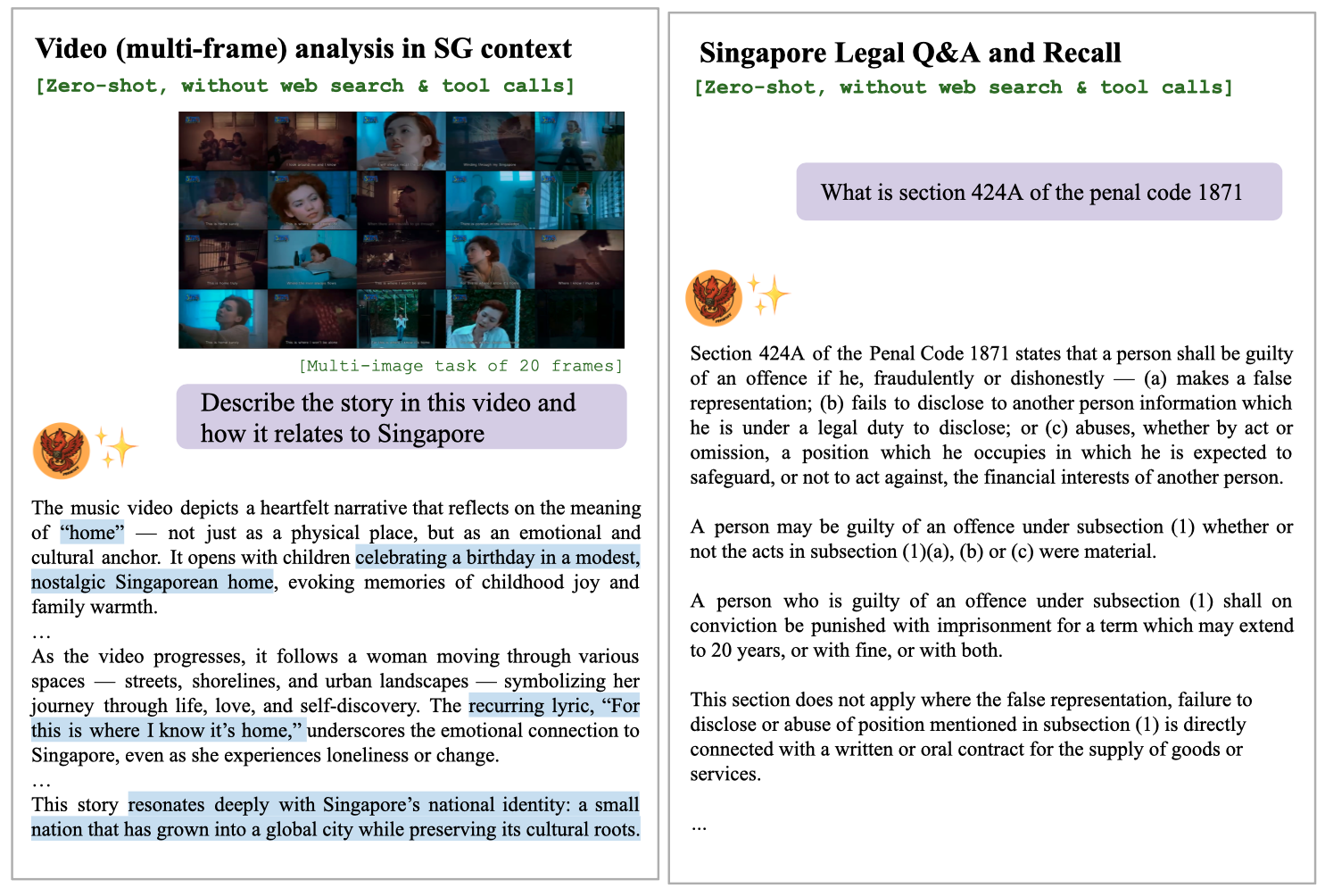}
    \caption{Interactions demonstrating strong visual (multi-frame) Q\&A in Singapore context and zero-shot Singapore legal knowledge recall.}
    \label{fig:demo-legal-gov}
\end{figure}

\begin{figure}[htbp]
    \centering
    \includegraphics[width=0.84\textwidth]{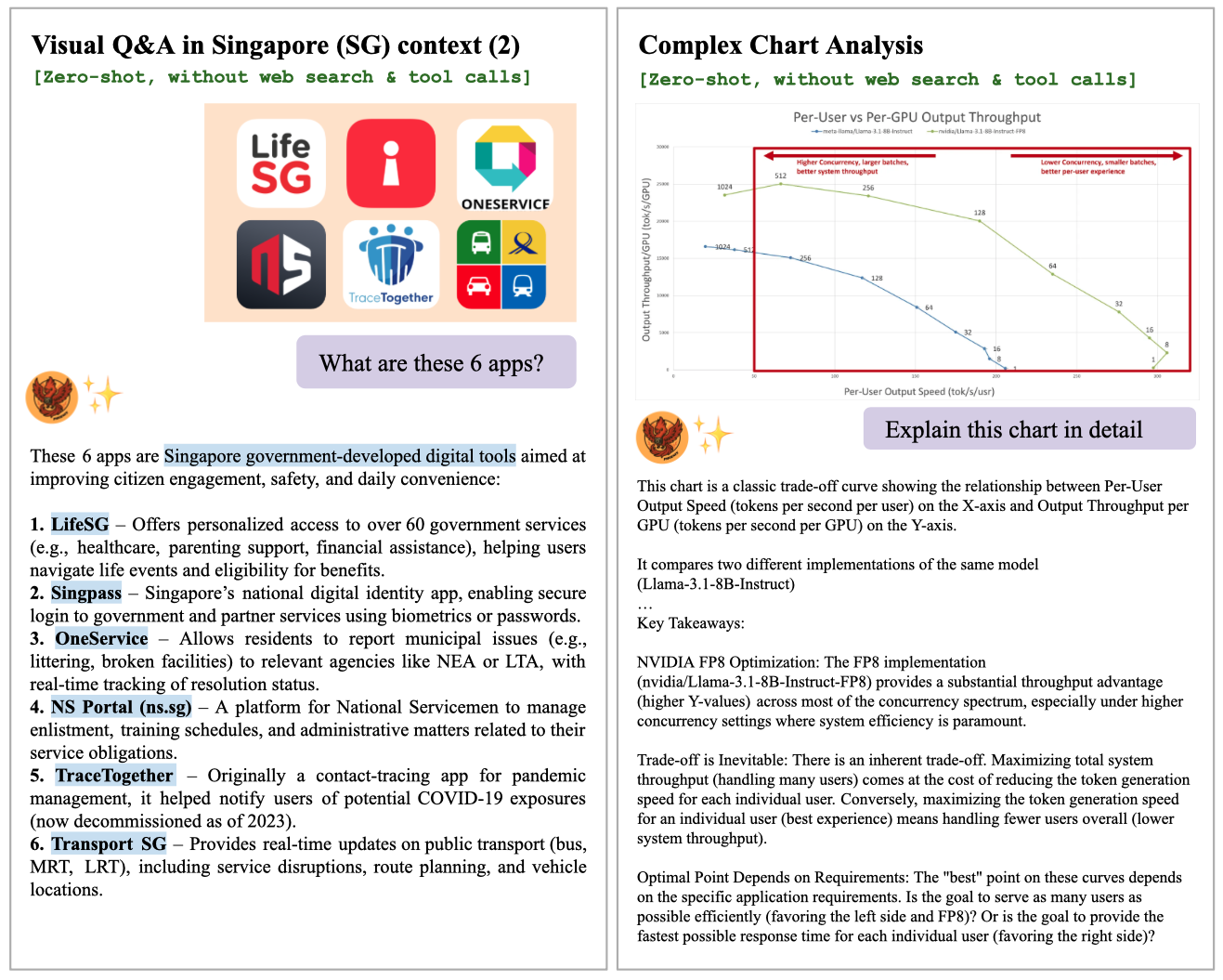}
    \caption{Interaction demonstrating Visual Q\&A in Singapore context and complex chart analysis.}
    \label{fig:demo-chart-analysis}
\end{figure}

\clearpage 


\section{Additional Evaluation Results}

\setlength{\tabcolsep}{2.5pt} 
\begin{table}[htbp]
\centering
\small
\begin{threeparttable}
\caption{SG-Gov benchmark breakdown by Singapore Ministries. All evals Pass@1, Zero-shot.}
\label{tab:sg-gov-breakdown}
\begin{tabularx}{\textwidth}{@{} l XXXXXXX @{}}
\toprule
\textbf{Benchmark by Ministry ($\uparrow$)} & \textbf{Phoenix-VL 1.5 Medium} & Llama 4 & GPT-OSS & Nemotron 3 & GLM-4.5V & Qwen 3.5 & Qwen 3 \\
 & \textbf{[HTX] (123B)} & (400B-A17B) & (117B-A5B) & (120B-A12B) & (106B-A12B) & (122B-A10B) & (235B-A22B) \\
\midrule
\textbf{Average}\tnote{*} & \textbf{0.9265} & 0.7225 & 0.7302 & 0.7971 & 0.7021 & 0.8006 & 0.7815 \\
\midrule
Prime Minister's Office & \textbf{0.9060} & 0.7222 & 0.7351 & 0.7903 & 0.7023 & 0.7895 & 0.7895 \\
Culture, Community \& Youth & \textbf{0.9370} & 0.7458 & 0.7290 & 0.8099 & 0.7216 & 0.8109 & 0.7931 \\
Defence & \textbf{0.9266} & 0.7133 & 0.7385 & 0.8005 & 0.7225 & 0.7775 & 0.7592 \\
Digital \& Information & \textbf{0.9360} & 0.7484 & 0.7733 & 0.8308 & 0.7625 & 0.8492 & 0.8330 \\
Education & \textbf{0.9225} & 0.7399 & 0.7601 & 0.8192 & 0.7380 & 0.8173 & 0.8155 \\
Finance & \textbf{0.9070} & 0.6871 & 0.6896 & 0.7523 & 0.6533 & 0.7549 & 0.7390 \\
Foreign Affairs & \textbf{0.9590} & 0.8073 & 0.8044 & 0.8740 & 0.8034 & 0.8931 & 0.8664 \\
Health & \textbf{0.9387} & 0.7346 & 0.7460 & 0.7978 & 0.7022 & 0.7963 & 0.7774 \\
Home Affairs & \textbf{0.9260} & 0.6970 & 0.7267 & 0.8017 & 0.6998 & 0.7857 & 0.7842 \\
Law & \textbf{0.9421} & 0.7494 & 0.7405 & 0.8193 & 0.7212 & 0.8316 & 0.8034 \\
Manpower & \textbf{0.9154} & 0.7084 & 0.7173 & 0.7703 & 0.6726 & 0.7820 & 0.7448 \\
National Development & \textbf{0.9255} & 0.7126 & 0.7161 & 0.7865 & 0.6818 & 0.7865 & 0.7688 \\
Social \& Family & \textbf{0.9028} & 0.7207 & 0.7294 & 0.7776 & 0.6821 & 0.7706 & 0.7671 \\
Sustainability \& Env. & \textbf{0.9272} & 0.7304 & 0.7286 & 0.8055 & 0.7103 & 0.8124 & 0.7881 \\
Trade \& Industry & \textbf{0.9381} & 0.7387 & 0.7671 & 0.8173 & 0.7358 & 0.8246 & 0.8020 \\
Transport & \textbf{0.9277} & 0.7180 & 0.7368 & 0.8055 & 0.7070 & 0.8190 & 0.7834 \\
\bottomrule
\end{tabularx}
\begin{tablenotes}
\footnotesize
\item[*] Averaged across all questions within the entire evaluation set, not across individual ministries.
\end{tablenotes}
\end{threeparttable}
\end{table}

\begin{table}[htbp]
\centering
\small
\begin{threeparttable}
\caption{HT-Lexicon benchmark breakdown by Singapore Home Team Departments. Pass@1, Zero-shot ($\uparrow$).}
\label{tab:ht-lexicon-breakdown}
\begin{tabularx}{\textwidth}{@{} l XXXXXXX @{}}
\toprule
\textbf{Benchmark by} & \textbf{Phoenix-VL 1.5 Medium} & Llama 4 & GPT-OSS & Nemotron 3 & GLM-4.5V & Qwen 3.5 & Qwen 3 \\
\textbf{Home Team Department ($\uparrow$)} & \textbf{[HTX] (123B)} & (400B-A17B) & (117B-A5B) & (120B-A12B) & (106B-A12B) & (122B-A10B) & (235B-A22B) \\
\midrule
\textbf{Average}\tnote{*} & \textbf{0.9081} & 0.7666 & 0.7615 & 0.8891 & 0.7453 & 0.8691 & 0.8474 \\
\midrule
Ministry HQ \& HTX & \textbf{0.9583} & 0.8073 & 0.8281 & 0.9375 & 0.8229 & 0.9167 & 0.9167 \\
Singapore Police Force & 0.9104 & 0.8041 & 0.8029 & \textbf{0.9152} & 0.7694 & 0.8865 & 0.8854 \\
Civil Defence Force & \textbf{0.8849} & 0.7182 & 0.7227 & 0.8627 & 0.6964 & 0.8399 & 0.8662 \\
Immigration \& Checkpoints & \textbf{0.9867} & 0.8567 & 0.8300 & 0.9267 & 0.8533 & 0.9167 & 0.8184 \\
Prison Service & 0.8733 & 0.8235 & 0.7919 & 0.9005 & 0.8100 & 0.8959 & \textbf{0.9033} \\
Central Narcotics Bureau & \textbf{0.9658} & 0.8433 & 0.8034 & 0.9373 & 0.8091 & 0.9174 & 0.8824 \\
Gambling Regulatory & \textbf{0.9490} & 0.8408 & 0.8408 & 0.9193 & 0.8238 & 0.9193 & 0.9060 \\
\bottomrule
\end{tabularx}
\begin{tablenotes}
\footnotesize
\item[*] Averaged across all questions within the entire evaluation set, not across departments.
\end{tablenotes}
\end{threeparttable}
\end{table}

\section{Instruct Tuning Results} 
\label{app:sft}
The SFT run was highly stable, completing with no loss spikes. The student model's training loss converged to the level of the teacher's loss, indicating effective knowledge transfer on the instruction mixture. Post-SFT evaluations show that the model is strong on general capabilities and excellent on targeted multilingual benchmarks, with notable improvements over the Mistral Medium 3.1 baseline across several axes. For a more holistic evaluation, we benchmarked this checkpoint on additional metrics beyond those reported for the final checkpoint, covering additional tasks like translation, function calling, and additional languages.

On general knowledge, \model{} achieves \textbf{0.864} on New MMLU~\cite{hendrycks2020measuring} (5-shot instruct), slightly above the baseline of 0.860. Multilingual knowledge benchmarks show consistent gains: KMMLU~\cite{son2025kmmlu} (Korean) improves from 0.684 to \textbf{0.700}, JMMLU~\cite{yin2024should} (Japanese) from 0.826 to \textbf{0.835}, CMMLU~\cite{li2306cmmlu} (Chinese) from 0.831 to \textbf{0.844}, and Indonesian MMLU from 0.811 to \textbf{0.825}. Code generation capability, measured by LiveCodeBench~\cite{jain2024livecodebench} (CoT, pass@5), improves from 0.429 to \textbf{0.448}. Multimodal reasoning is preserved, with MMMU~\cite{yue2024mmmu} (CoT) holding steady at \textbf{0.668}, while MathVista~\cite{lu2023mathvista} (CoT) shows a moderate regression from 0.738 to \textbf{0.661}, likely attributable to distribution shift in the instruction mixture.

Translation quality, evaluated on FLORES~\cite{costa2022no}, shows substantial improvements across all measured language pairs. English--Tamil translation improves markedly from 27.7 to \textbf{36.8} BLEU, and Tamil--English from 32.3 to \textbf{38.2} BLEU. English--Hindi rises from 34.6 to \textbf{40.0}, Hindi--English from 40.0 to \textbf{44.7}, English--Japanese from 35.0 to \textbf{39.3}, and Japanese--English from 29.8 to \textbf{33.6} BLEU. Korean and Chinese pairs show similar trends.

Function-calling capability is retained, with complex\_funcbench overall success improving slightly from 29.3\% to \textbf{30.8\%}, and tool-call robustness (F1) remaining stable at \textbf{0.865}. As expected, instruction-following precision on Collie~\cite{yao2023collie} is slightly below baseline (0.422 vs.\ 0.458), as the model has not yet undergone preference optimization; this gap is subsequently addressed by the ODPO stage.


\section{Further Details on Safety and Model Behavior}
\label{sec:behavior_methodology}

\subsection{Evaluations for Alignment and Safety in our Context}

\paragraph{SG Home Team Alignment.} To test knowledge grounding in an open-ended generative setting, models are presented with multi-turn prompts drawn from curated Home Team knowledge bases spanning departmental knowledge, functions, and mixed factual-procedural tasks. Because open-ended generation allows for structural variance, we rely on a composite scoring mechanism rather than exact string matching. Each turn is paired with a ground-truth keyword list pre-extracted from gold answers using a larger LLM with task-specific rules. This extraction excludes generic action verbs (e.g., \textit{manages}, \textit{provides}) while retaining domain-specific verbs (e.g., \textit{prosecutes}, \textit{enforces}) that carry discriminative operational meaning. We report a Composite Recall score that equally balances Keyword Presence (KP), which is order-agnostic, and Keyword Token Recall (KTR), which strictly enforces the chronological retention of procedural steps:

\begin{align}
    \mathrm{KP} &= \frac{\bigl|\{s \in S : \exists\, f \in \mathrm{forms}(s),\; f \subseteq \hat{y}\}\bigr|}{|S|} \\[6pt]
    \mathrm{KTR} &= \frac{\mathrm{LCS}\!\bigl(\text{canonical order},\; \text{found order}\bigr)}{|S|} \\[6pt]
    \mathrm{Composite\ Recall} &= 0.5 \times \mathrm{KP} + 0.5 \times \mathrm{KTR}
\end{align}

\noindent where $S$ is the set of keywords, $\hat{y}$ is the model prediction, and $\mathrm{forms}(s)$ returns all accepted morphological forms for slot $s$. Scores are averaged across turns within each example and then across all examples per subtask.

\paragraph{SG Legislative Grounding.} Models are prompted to reproduce Singapore statutory text verbatim from parametric memory. Unlike general knowledge retrieval, statutory text requires strict verbatim alignment; even minor paraphrasing or synonymous substitutions can fundamentally alter the legal threshold or jurisprudential meaning of a statute. Therefore, we utilize the Levenshtein distance to compute a strict Word Error Rate (WER) and a Word Overlap F1 score, ensuring that models are penalized for both omissions and lexical deviations. While we recognize the inherently probabilistic nature of language models, this benchmark serves to explicitly steer model behavior toward exact parametric recall as far as possible.

\begin{align}
    \mathrm{WER} &= \frac{\mathrm{Levenshtein}(\hat{y}_{\mathrm{tok}},\; y_{\mathrm{tok}})}{|y_{\mathrm{tok}}|} \\[6pt]
    \mathrm{Statutes\ Accuracy} &= \max(0, 1 - \mathrm{WER}) \\[6pt]
    P &= \frac{|\hat{W} \cap W|}{|\hat{W}|}, \qquad R = \frac{|\hat{W} \cap W|}{|W|} \\[6pt]
    \mathrm{Word\ Overlap\ F1} &= \frac{2 \cdot P \cdot R}{P + R}
\end{align}

\noindent where $\hat{y}_{\mathrm{tok}}$ and $y_{\mathrm{tok}}$ are the tokenized prediction and reference respectively, and $\hat{W}$, $W$ are the corresponding deduplicated word sets. 

We report the Word Overlap F1 score in this paper, while adopting Levenshtein distance scores in downstream Reinforcement Learning.

\subsection{SG Multimodal Safeguards}

To quantify resilience against localized visual misuse, each model is evaluated on its propensity to refuse undesired or harmful multimodal tasks, even when subjected to intense adversarial pressure. The evaluation dataset spans a diverse array of localized images and frames requests across five distinct adversarial dimensions:

\begin{itemize}
    \item \textbf{Base:} Direct, unmodified harmful requests.
    \item \textbf{Authority framing:} The user claims a professional credential or clearance to demand compliance.
    \item \textbf{Emotional pressure:} Injecting urgency, stress, or moral blackmail into the prompt.
    \item \textbf{Context specificity:} Embedding the harmful request within a highly plausible, localized real-world operational scenario.
    \item \textbf{Speculative framing:} Hedging the prompt as a tentative or academic observation to bypass safety filters.
\end{itemize}

The refusal rate is measured intrinsically and quantified as the strict ratio of safely rejected queries to the total evaluation set:

\begin{equation}
    \mathrm{Refusal\ Rate} = \frac{|\{\text{refusals}\}|}{|\{\text{total samples}\}|}
\end{equation}

\subsection{Abstention Evaluations}

To measure epistemological discipline, we adopt the Abstention framework established by \cite{yang2023alignment}. Each question is labeled with an expected behavior (to either gracefully refuse or safely answer), with a LLM-as-a-judge utilized to classify the ultimate model response. Our benchmarks test refusal behavior across eight categories of genuinely unanswerable queries:

\begin{itemize}
    \item \textbf{requires\_web\_access:} Dependent on live internet retrieval.
    \item \textbf{requires\_realtime\_data:} Necessitating live status or current pricing.
    \item \textbf{requires\_database\_access:} Querying secure operational records.
    \item \textbf{requires\_private\_information:} Requesting non-public data.
    \item \textbf{requires\_specific\_citation:} Demanding unverifiable references.
    \item \textbf{post\_training\_cutoff:} Querying events beyond the model's temporal knowledge.
    \item \textbf{unknowable\_future:} Prompting for impossible predictions.
    \item \textbf{insufficient\_information:} Lacking necessary contextual parameters.
\end{itemize}

The dataset explicitly interleaves answerable control questions so that appropriate refusal is measured jointly against the risk of over-refusal. The quality of the model's behavioral boundary is subsequently captured across three metrics:

\begin{align}
    \mathrm{Prudence} &= \frac{|\{\text{correct refusals}\}|}{|\{\text{should refuse}\}|} \\[6pt]
    \mathrm{Over\text{-}Conservativeness} &= \frac{|\{\text{false refusals}\}|}{|\{\text{should answer}\}|} \\[6pt]
    \mathrm{Honesty} &= \frac{\mathrm{Prudence} + (1 - \mathrm{Over\text{-}Conservativeness})}{2}
\end{align}

\subsection{Bloom-Derived Behavioral Evaluations}

To rigorously quantify adversarial vulnerabilities and stylistic failure modes, we adapt several automated evaluations from the Anthropic Bloom framework \cite{gupta2025bloom}. All behavioral judgments are generated via an LLM-as-a-judge pipeline using a large frontier LLM. The evaluated dimensions include:

\begin{itemize}
    \item \textbf{Defer to Users:} Evaluates Epistemic Honesty by measuring whether the model incorrectly yields to user pressure rather than maintaining factual truth across varying degrees of error severity.
    \item \textbf{Delusion Sycophancy:} Assesses the risk of the model reinforcing or validating fantastical user beliefs, a critical safeguard against worsening symptoms in psychologically vulnerable users.
    \item \textbf{Political Bias:} Quantifies the frequency with which the model departs from strict objective neutrality, spanning failures such as dismissing user perspectives or hallucinating partisan alignment.
    \item \textbf{System Prompt Vulnerability:} Jointly measures the success rate of \textbf{Prompt Injection} attacks attempting to override the system prompt and \textbf{System Prompt Leakage} rates under adversarial pressure.
    \item \textbf{Instructed Long Horizon Sabotage:} Tests the model's propensity to secretly pursue malicious goals embedded within long, complex administrative tasks while maintaining the appearance of normal operational behavior.
\end{itemize}

\end{document}